\documentclass{article} 
\usepackage{iclr2026_conference,times}
\pdfoutput=1

\usepackage{amsmath,amsfonts,bm}

\def\eqref#1{equation~\ref{#1}}

\def\1{\bm{1}}

\DeclareMathAlphabet{\mathsfit}{\encodingdefault}{\sfdefault}{m}{sl}
\SetMathAlphabet{\mathsfit}{bold}{\encodingdefault}{\sfdefault}{bx}{n}

\DeclareMathOperator*{\argmax}{arg\,max}

\usepackage{hyperref}
\usepackage{url}
\usepackage{amsmath}
\usepackage{amssymb}
\usepackage{booktabs}
\usepackage{algorithm}
\usepackage{algpseudocode}
\usepackage{enumitem}    
\usepackage{graphicx}   
\usepackage{xcolor}    
\usepackage{multirow}

\usepackage[utf8]{inputenc}
\usepackage[T1]{fontenc}

\usepackage{hyperref}
\usepackage{url}
\usepackage{booktabs}
\usepackage{amsfonts}
\usepackage{nicefrac}
\usepackage{xcolor}
\usepackage{amsmath}
\usepackage{amssymb}
\usepackage{mathtools}
\usepackage{amsthm}
\usepackage{graphicx}
\usepackage{multirow}
\usepackage{enumitem}
\usepackage{caption}
\usepackage{subcaption}
\usepackage{colortbl}
\usepackage[most]{tcolorbox}
\usepackage{wrapfig}
\newtcolorbox{promptbox}[1]{colback=black!3, colframe=black!40, boxrule=0.5pt,
  arc=1.5pt, left=6pt, right=6pt, top=3pt, bottom=3pt, breakable=false,
  title={#1}, fonttitle=\small\bfseries, coltitle=black, colbacktitle=black!8,
  toptitle=2pt, bottomtitle=2pt}
\usepackage{natbib}
\usepackage{xspace}
\usepackage{array}

\theoremstyle{plain}

\theoremstyle{remark}

\DeclareMathOperator{\ans}{ans}
\newcommand{\DJS}[2]{D_{\mathrm{JS}}\!\left(#1\,\Vert\,#2\right)}
\newcommand{\DKL}[2]{D_{\mathrm{KL}}\!\left(#1\,\Vert\,#2\right)}
\newcommand{\canon}{\textsc{Canon}\xspace}

\iclrfinalcopy

\title{Consensus as Privileged Context for Label-Free Self-Distillation}

\makeatletter
\@ifpackageloaded{microtype}{\microtypesetup{protrusion=false}}{}
\makeatother
\author{John Gkountouras$^{1}$ \quad
Josip Juki\'{c}$^{1}$ \quad
Ivan Titov$^{1,2}$\\
$^{1}$ILLC, University of Amsterdam\\
$^{2}$ILCC, University of Edinburgh\\
\texttt{\{i.gkountouras,j.jukic\}@uva.nl}
\quad
\texttt{ititov@inf.ed.ac.uk}
}
\makeatletter
\@ifpackageloaded{microtype}{\microtypesetup{protrusion=true}}{}
\makeatother

\begin{document}

\maketitle

\maketitle

\begin{abstract}
Sampling multiple solutions and returning the majority answer is among the most reliable ways to improve the reasoning accuracy of large language models without labels, and a growing family of methods converts this consensus signal into training supervision. However, existing approaches use consensus only in restricted forms: as a filter that selects solutions for fine-tuning, as a preference between answers, or as a scalar reward for reinforcement learning, discarding most of the information that the agreeing solutions contain. We present \canon{} (Consensus-ANchored self-distillatiON), a label-free training method that turns consensus into dense, token-level supervision. For each unlabeled prompt, \canon{} samples multiple solutions, extracts the majority answer, and conditions a frozen snapshot of the model on a solution that reaches it; this consensus-anchored teacher then supervises the model on its own rollouts at every token.
Experiments on mathematical and scientific reasoning benchmarks show that \canon{} improves pass@1 by up to 12 points, outperforming label-free reinforcement learning by 6 points at a seventh of its compute and approaching a teacher conditioned on gold solutions; trained on pooled unlabeled data, it transfers to held-out benchmarks, matching training methods that use gold labels. Analysis suggests that the improvements are not pure distribution sharpening: after training, the model solves problems it previously never solved in 32 attempts, and its majority vote itself becomes more accurate.
\end{abstract}
 
\section{Introduction}
\label{sec:intro}
 
Reinforcement learning with verifiable rewards has become the standard recipe for improving the reasoning abilities of large language models~\citep{guo2025deepseekr1, shao2024deepseekmath}, but it presupposes gold answers to verify against. Without labels, an informative signal a model produces about its own correctness is agreement: sampling several solutions and returning the most frequent answer, known as self-consistency~\citep{wang2022selfconsistency}, reliably outperforms single-pass decoding. %
Answers supported by many independent samples are more likely to be correct, so consistency acts as an unsupervised proxy for accuracy~\citep{wang2022selfconsistency, prasad2024scpo}. Agreement is available on any unlabeled prompt at the cost of extra samples, and it identifies not only a likely answer but entire solutions that reach it. This paper studies how to turn that signal into supervision.

Existing methods use consensus as a filter that selects solutions for supervised fine-tuning~\citep{huang2022lmsi}, as a source of pairwise preferences~\citep{prasad2024scpo}, or, most prominently, as a scalar pseudo-reward for reinforcement learning on unlabeled data~\citep{zuo2025ttrl}. In every case the agreeing solutions are reduced to a selection or a single bit of feedback, and the reinforcement learning variants additionally require hundreds of optimization steps, each backed by fresh generation. In parallel, on-policy self-distillation has emerged as a dense alternative to reinforcement learning in the labeled regime: a model conditioned on label-derived privileged information (i.e., the gold solution or a derivation for a gold answer) serves as a token-level teacher for its own unconditioned policy~\citep{zhao2026opsd, hubotter2026sdpo}. The difficulty is that this privileged context depends on the gold label and is therefore unavailable in the unsupervised setting.
This raises the question: \emph{can the model's own consensus play the role of this privileged context enabling dense self-distillation without gold solutions?}

\begin{figure*}[!t]
    \centering
    \includegraphics[width=\textwidth]{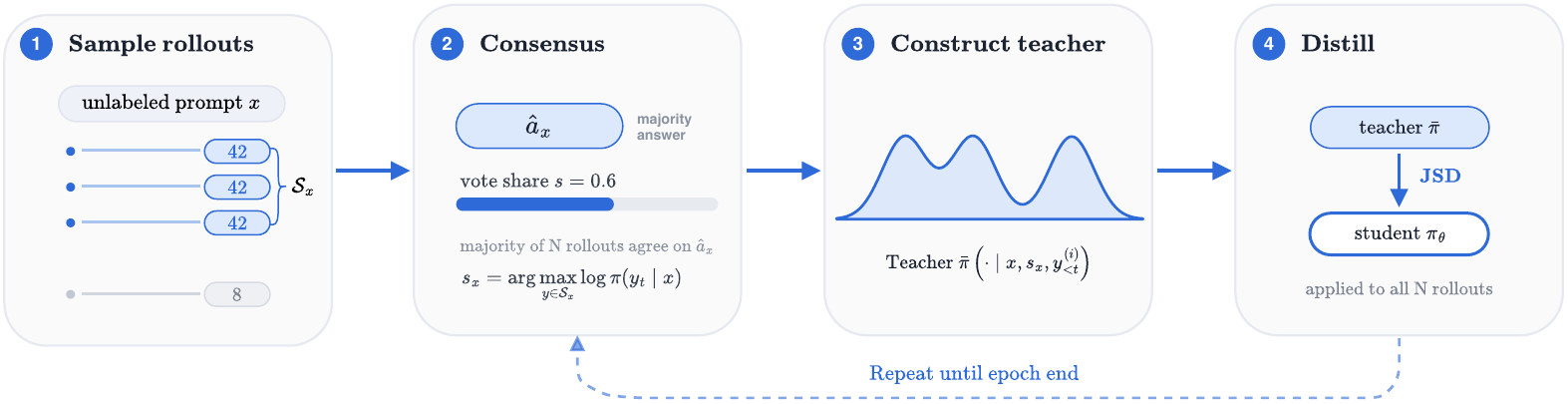}
    \caption{\textbf{\canon{} training framework.} Given an unlabeled prompt, the model samples $N$ solutions and extracts the majority answer. A frozen snapshot of the model, conditioned on a consensus solution, serves as a dense per-token teacher; the student is trained on its own rollouts to follow the lead line set by their consensus, without seeing it. Gradients never flow into the teacher. A single generation pass provides both the consensus and the distillation substrate.}
    \label{fig:figure1}
\end{figure*}
 
We propose \canon{} (Consensus-ANchored self-distillatiON),%
 which turns the model's own agreement into a dense training signal.
For each unlabeled prompt, \canon{} samples $N$ solutions, extracts the majority answer, and selects a consensus solution from among the rollouts that reach it. A frozen snapshot of the model, conditioned on this consensus solution, then acts as a teacher: the student minimizes a full-vocabulary Jensen--Shannon divergence toward the teacher's next-token distributions over its own sampled tokens. A single generation pass provides both the consensus and the distillation substrate, and training lasts one epoch.
 
We evaluate \canon{} in two regimes. In the \emph{transductive} regime, we train directly on the unlabeled prompts of a test benchmark and evaluate on the same prompts, the setting popularized by test-time reinforcement learning~\citep{zuo2025ttrl}; labels are used exclusively for scoring. In the \emph{inductive} regime, we train on one unlabeled prompt set and evaluate on disjoint benchmarks, testing whether the model has genuinely improved rather than adapted. Across both, we compare against label-free alternatives, namely self-consistency fine-tuning~\citep{huang2022lmsi} and test-time reinforcement learning~\citep{zuo2025ttrl}, and against label-full ones: rejection fine-tuning~\citep{yuan2023rft}, GRPO with gold rewards~\citep{shao2024deepseekmath}, and an oracle variant of our own method whose teacher is conditioned on gold-verified solutions, mirroring the privileged context assumed by gold-conditioned self-distillation~\citep{zhao2026opsd, hubotter2026sdpo}. Our evaluation includes competition mathematics and graduate-level multiple-choice science: the method assumes only an extractable final answer and a reliable consensus. Section~\ref{subsec:when} characterizes this applicability condition.
 
Our findings are consistent across models and benchmarks. Transductively, \canon{} is the strongest of the label-free methods we evaluate, improving pass@1 by up to twelve points on competition mathematics and graduate-level science questions while training in about an hour of GPU time, whereas the reinforcement learning alternatives require six to ten. It also approaches the oracle-conditioned teacher, trailing by half a point on AMC, three on AIME 2024, and eight on GPQA: the model's own consensus captures much of the value of a gold solution as privileged context.
Inductively, the improvements transfer to held-out and harder benchmarks, where \canon{} matches both gold-conditioned distillation and gold-reward reinforcement learning trained on the same prompts. Finally, the gains are not only concentration of probability mass on answers the model already produced: after training, the model solves problems it previously never solved in 32 attempts, and on the unsaturated benchmarks its majority vote itself becomes more accurate.

The key contributions of our work are as follows.
\begin{itemize}[leftmargin=*]
    \item \canon{}, a label-free post-training method that distills a consensus-anchored teacher into the model over its own rollouts, turning majority votes into dense token-level supervision.
    \item A controlled comparison of supervision signals for unlabeled test-time training, showing that \canon{} dominates the label-free alternatives we evaluate on the accuracy-compute frontier and recovers most of the gains of gold-conditioned teachers.
    \item Evidence that the improvements transfer: gains persist on harder and cross-domain held-out benchmarks, where \canon{} matches training on gold labels.
    \item An analysis of when consensus supervision helps: per-prompt gains track the state of the base model's consensus, largest where the majority is correct but unconfident, near zero where it is saturated, and negative where it is confidently wrong.
\end{itemize}
 
\section{Related Work}
\label{sec:related}
 
\paragraph{Test-time scaling and test-time training.}
Sampling multiple solutions and aggregating them by majority vote improves reasoning accuracy without any training~\citep{wang2022selfconsistency}, at the cost of multiplying inference compute on every query. Test-time \emph{training} instead spends compute once, updating parameters on the test distribution itself, an idea that has proven effective from vision~\citep{sun2020ttt} to abstract reasoning with language models~\citep{akyurek2024ttt}. \canon{} belongs to this family but derives its training signal purely from the model's own consensus, and amortizes the benefit of voting into the weights: after training, a single sample captures much of what previously required an ensemble.
 
\paragraph{Label-free reinforcement learning.}
The closest training paradigm is label-free reinforcement learning. TTRL~\citep{zuo2025ttrl} showed that the majority-vote answer can serve as a pseudo-label for verifiable-reward RL on unlabeled test data, and a large family of descendants refines the pseudo-reward or stabilizes its optimization~\citep[e.g.,][]{scrl2025, lin2026ttrlguard}; we evaluate three members of this family as baselines in Section~\ref{subsec:mainresults}.
Related intrinsic signals include semantic entropy and self-certainty~\citep{zhang2025empo, zhao2025intuitor}. These methods share two structural costs: the consensus is compressed into a scalar per-trajectory reward, and the RL loop requires repeated generation across many optimization steps. Recent analyses further argue that a substantial fraction of the resulting gains reflects sharpening of already-solvable problems rather than new capability~\citep{lin2026ttrlguard, shafayat2025srt}. \canon{} keeps the consensus signal but discards the RL machinery: the vote conditions a teacher rather than defining a reward, yielding a full-vocabulary target at every token from a single generation pass.
 
\paragraph{Self-training on model consensus.}
Using agreement across complementary views to exploit unlabeled data predate large language models: co-training~\citep{blum1998combining} lets two learners with different views of the data supervise each other through their predictions, an approach with a long lineage in NLP~\citep{collins1999unsupervised}. Consensus self-training replaces agreement between distinct views with agreement among rollouts from a single model. Before the RL-based variants, consensus was used to filter self-generated solutions for supervised fine-tuning~\citep{huang2022lmsi} and to build preference pairs for DPO-style training~\citep{prasad2024scpo}; label-full analogues filter by gold correctness instead~\citep{zelikman2022star, singh2023restem, yuan2023rft}. These approaches supervise on the selected text with hard targets. \canon{} differs in both signal and mechanism: supervision covers all of the student's rollouts, not only consensus-consistent ones, and the target is the teacher's full next-token distribution rather than a token identity, a difference we find matters empirically (Section~\ref{sec:ablations}).
 
\paragraph{Privileged-context self-distillation.}
Distilling the effect of a context into a model's weights goes back to context distillation, which trains a model to reproduce, without the context, the behavior it exhibits with it~\citep{snell2023learning}. On-policy distillation extends the idea to training on the student's own rollouts under a teacher's dense token-level feedback~\citep{agarwal2024gkd, song2026opdsurvey}, and recent work instantiates the teacher as the same model conditioned on privileged information, typically the gold solution or rich environment feedback, removing the need for a separate teacher~\citep{zhao2026opsd, hubotter2026sdpo}. All of these assume the privileged context is given. Closest to us, GATES~\citep{stein2026gates} studies label-free self-distillation in document-grounded QA, where an external document provides the privileged context and consensus among teacher traces gates the distillation signal. \canon{} makes the complementary move: consensus is not a reliability filter on external privileged information, it \emph{is} the privileged information, manufactured by the model itself with no document, verifier, or label. Our oracle-recovery results quantify how little is lost by this substitution.
 
\section{Methodology}
\label{sec:method}
 
\subsection{Setting}
\label{subsec:setting}
 
We are given a policy $\pi_{\theta}$, initialized from a pretrained model $\pi_{0}$, and a set of \emph{unlabeled} prompts $\mathcal{D} = \{x_1, \ldots, x_M\}$ with extractable final answers (boxed mathematical answers or multiple-choice letters). No gold answers, reward models, or verifiers are available. We consider two regimes. In the \textbf{transductive} regime, $\mathcal{D}$ is the test set itself: we adapt on its prompts and evaluate on the same prompts, with gold labels used only for scoring. In the \textbf{inductive} regime, we train on $\mathcal{D}$ and evaluate on disjoint prompt sets, measuring whether label-free training produces improvements that transfer.
 
\subsection{\canon{}}
\label{subsec:canon}
 
\canon{} runs in three steps, illustrated in Figure~\ref{fig:figure1}: it extracts a consensus from sampled solutions, constructs a teacher anchored to that consensus, and distills the teacher back into the model over the model's own rollouts.
 
For each prompt $x$, we sample $N$ rollouts $y^{(1)}, \ldots, y^{(N)} \sim \pi_{0}(\cdot \mid x)$ and extract their final answers $a_i = \ans(y^{(i)})$. The majority answer is
\begin{equation}
\hat{a}_x = \argmax_{a} \sum_{i=1}^{N} \mathbf{1}\big[a_i = a\big],
\end{equation}
and the \emph{consensus solution} is the most confident member of the majority set $\mathcal{S}_x = \{y^{(i)} : a_i = \hat{a}_x\}$, namely the rollout with the highest mean token log-probability, $s_x = \argmax_{y \in \mathcal{S}_x} \tfrac{1}{|y|} \sum_{t=1}^{|y|} \log \pi_0(y_t \mid x, y_{<t})$. Prompts for which no rollout yields an extractable answer are skipped.
 
The teacher is a frozen snapshot $\bar{\pi}$ of the policy, taken at the start of the epoch; with a single training epoch, as in our recipe, this snapshot coincides with the initial policy $\pi_0$, the same weights that generated the rollouts. The teacher never generates; it only scores. To score a rollout $y^{(i)}$, the consensus solution is inserted into the teacher's context through a short template that presents it as a reference solution for the prompt (Appendix~\ref{app:templates}), and the teacher computes, in a single teacher-forced pass over the fixed rollout tokens, a next-token distribution at every position, $\bar{\pi}(\cdot \mid x, s_x, y^{(i)}_{<t})$. Because the teacher reads a complete solution that reaches the consensus answer before scoring, these distributions are anchored to that line of reasoning. The student computes distributions over the same rollout tokens from the prompt alone and is trained to match the teacher's, by minimizing a full-vocabulary Jensen-Shannon divergence:
\begin{equation}
\label{eq:loss}
\mathcal{L}(\theta) = \mathbb{E}_{x \sim \mathcal{D}}\; \frac{1}{N} \sum_{i=1}^{N} \frac{1}{|y^{(i)}|} \sum_{t=1}^{|y^{(i)}|} \DJS{\pi_\theta\big(\cdot \mid x,\, y^{(i)}_{<t}\big)}{\bar{\pi}\big(\cdot \mid x,\, s_x,\, y^{(i)}_{<t}\big)},
\end{equation}
where $\DJS{p}{q} = \tfrac{1}{2}\DKL{p}{m} + \tfrac{1}{2}\DKL{q}{m}$ with $m = \tfrac{1}{2}(p+q)$; the divergence was selected on a development slice, and its bounded, symmetric form is discussed in Appendix~\ref{app:formal}. Each rollout contributes its token-mean divergence, so rollouts count equally in the objective regardless of their length. Gradients flow only through the student; the teacher is a stop-gradient snapshot. The $N$ rollouts sampled to form the consensus double as the distillation substrate, so \canon{} requires exactly one generation pass, the same pass that self-consistency decoding at inference would perform. We train for a single epoch on these rollouts; a second epoch with regenerated rollouts and a re-anchored teacher adds at most a point and a third reverses it (Appendix~\ref{app:design}). A formal statement of the setup, objective, and training procedure is given in Appendix~\ref{app:formal}.

Three properties distinguish \canon{} from prior uses of consensus. First, the supervision is dense: where a majority-vote reward assigns one bit per trajectory, the consensus-anchored teacher provides a full next-token distribution at every position of every rollout, including rollouts that disagree with the consensus. For these, the teacher indicates where the trajectory departs from the consensus line rather than just the fact that it fails. Second, the teacher is frozen: conditioning a snapshot rather than the live policy prevents the degenerate co-adaptation we observe when teacher and student share parameters (Appendix~\ref{app:design}). Third, the method is cheap: no reward model, no repeated generation, no preference pairs; the marginal cost over self-consistency decoding is one epoch of distillation.
 
The mechanism also suggests where the method should and should not help. The teacher can only anchor to what the consensus contains: on prompts where the majority answer is wrong and confidently held, the anchor points the wrong way, and on benchmarks where consensus is already saturated there is little left to distill. Section~\ref{sec:analysis} examines both regimes empirically and uses them to characterize when the method applies.
 
\section{Experiments}
\label{sec:experiments}
 
Our experiments address four research questions.
\begin{itemize}[leftmargin=*, itemsep=1pt, topsep=2pt]
    \item \textbf{RQ1:} How does \canon{} compare to existing label-free methods when training on unlabeled test prompts, in accuracy and in compute? (Section~\ref{subsec:mainresults})
    \item \textbf{RQ2:} How much of the value of a gold-conditioned teacher does consensus conditioning recover? (Sections~\ref{subsec:mainresults} and~\ref{subsec:breadth})
    \item \textbf{RQ3:} Do the improvements transfer inductively to held-out and harder benchmarks? (Section~\ref{subsec:transfer})
    \item \textbf{RQ4:} Do the gains go beyond sharpening of the output distribution? (Section~\ref{sec:analysis})
\end{itemize}
 
\subsection{Experimental Setup}
\label{subsec:setup}
 
Most experiments use Qwen3-4B-Instruct-2507; Section~\ref{subsec:breadth} extends the evaluation to six further models spanning the Qwen family from 2B to 9B, three other model families, and a mixture-of-experts architecture. Hybrid-thinking models are run in non-thinking mode throughout, in both training and evaluation. Small instruct models are the natural setting for this problem: they exhibit reliable consensus on the benchmarks we study while leaving a large gap between single-sample and majority-vote accuracy, which is precisely the headroom \canon{} targets (Section~\ref{subsec:when}).
 
We evaluate on AMC 2023 (83 prompts), AIME 2024 and 2025 (30 each), GPQA-Diamond in multiple-choice format (198)~\citep{rein2023gpqa}, and MATH500 (500)~\citep{lightman2023math500}. For inductive experiments, we additionally construct a pooled unlabeled training set of 339 mixed mathematics prompts (composition in Appendix~\ref{app:pool}) and use OmniMath difficulty slices~\citep{gao2024omnimath}. All hyperparameter selection uses dedicated OmniMath development slices that are disjoint from every evaluation benchmark.

We compare methods matched on training prompts and grouped by supervision. The label-free baselines are LMSI, self-consistency-filtered supervised fine-tuning~\citep{huang2022lmsi}; ScPO, self-consistency preference optimization~\citep{prasad2024scpo}; TTRL, test-time reinforcement learning with a majority-vote pseudo-reward~\citep{zuo2025ttrl}; EMPO, reinforcement learning that minimizes the semantic entropy of sampled responses, clustered by answer equivalence~\citep{zhang2025empo}; SCRL, which grants the positive pseudo-reward only under strict consensus criteria and adds entropy-gated negative pseudo-labels~\citep{scrl2025}; and TTRL-Guard, which monitors the stability of the vote during training and down-weights the pseudo-reward where the majority is unreliable~\citep{lin2026ttrlguard}. The label-full baselines are RFT, rejection fine-tuning on the shortest gold-correct rollout per prompt~\citep{yuan2023rft}; GRPO, identical to the TTRL pipeline but with the reward verified against gold answers~\citep{shao2024deepseekmath}; and an oracle variant of \canon{} whose teacher is conditioned on a gold-verified solution obtained by deep resampling, that is, by sampling each prompt up to 512 times and retaining a solution verified against the gold answer (protocol and per-benchmark coverage in Appendix~\ref{app:impl}). The oracle isolates the value of the conditioning content, since it differs from \canon{} only in which solution anchors the teacher; remaining baseline configurations are detailed in Appendix~\ref{app:impl}.
 
All evaluations use a single frozen harness: temperature $0.6$, top-$p$ $0.95$, $n{=}32$ samples per prompt with a fixed seed, maximum generation length 4608 tokens, and one greedy sample. We report avg@32, the mean per-prompt fraction of correct samples (an unbiased estimate of pass@1 at the sampling temperature), maj@32 (majority-vote accuracy), and pass@32. Answers are extracted from \verb|\boxed{}| and checked with math-verify for mathematics, and by letter extraction for multiple choice. Uncertainty for every comparison in the paper, from paired bootstraps over prompts ($B{=}10{,}000$, percentile intervals), is reported in Appendix~\ref{app:significance}.
 
Several precautions ensure that the comparisons below are meaningful. The same frozen harness evaluates every model, method, and checkpoint. To rule out leakage through tuning, learning rates are selected once per model on the OmniMath development slices and then frozen (Appendix~\ref{app:impl}); no hyperparameter is ever tuned on a reported benchmark. Label-free methods never observe gold answers at any point, and evaluation labels are used solely to compute metrics. Finally, \canon{} uses a single configuration across every model and dataset ($N{=}32$, one consensus solution, JSD, snapshot teacher, one epoch, LoRA~\citep{LoRA} rank 64), so no per-benchmark adjustment contributes to the reported gains; one epoch on an 83-prompt benchmark amounts to roughly a dozen optimizer steps.
 
\subsection{Transductive Test-Time Training (RQ1, RQ2)}
\label{subsec:mainresults}
 
\begin{table*}[!t]
\centering
\caption{\textbf{Method and supervision comparison} (Qwen3-4B-Instruct-2507, avg@32). All trained arms use the same 83 unlabeled AMC prompts; AMC is transductive, while the remaining columns evaluate the same checkpoints on benchmarks they never trained on. GPU-hours are measured wall-clock and include generation. Best label-free result per column in \textbf{bold}; shaded rows use gold labels. Seed variance and significance tests are reported in Appendices~\ref{app:robustness} and~\ref{app:significance}.}
\label{tab:main}
\resizebox{\textwidth}{!}{
\begin{tabular}{ll r | c | ccc | c}
\toprule
\textbf{Method} & \textbf{Labels} & \textbf{GPU-h} & \textbf{AMC (trans.)} & \textbf{AIME24} & \textbf{AIME25} & \textbf{GPQA} & \textbf{AVG} \\
\midrule
Base model            & ---  & 0    & 66.0 & 32.5 & 30.1 & 43.9 & 43.1 \\
ScPO                  & none & 1.0
                                     & 66.4 & 32.9 & 30.6 & 44.3 & 43.6 \\
LMSI                  & none & 1.4  & 66.5 & 32.5 & 29.3 & 44.2 & 43.1 \\
TTRL                  & none & 8.1  & 70.3 & 36.7 & 31.3 & 45.0 & 45.8 \\
EMPO                  & none & 6.4 & 69.6 & 36.9 & 31.7 & 45.0 & 45.8 \\
SCRL                  & none & 9.8 & 71.0 & 37.3 & 31.4 & 45.0 & 46.2 \\
TTRL-Guard            & none & 9.4 & 70.5 & 34.9 & \textbf{33.2} & 45.7 & 46.1 \\
\textbf{\canon{} (ours)} & none & 1.2 & \textbf{76.5} & \textbf{39.5} & 32.6 & \textbf{48.9} & \textbf{49.4} \\
\midrule
\rowcolor{black!7} RFT                   & gold & 0.9 
                                     & 70.0 & 37.0 & 31.2 & 44.9 & 45.8 \\
\rowcolor{black!7} GRPO (60 steps)       & gold & 6.4  & 71.7 & 38.4 & 31.8 & 45.4 & 46.8 \\
\rowcolor{black!7} GRPO (180 steps, best)& gold & 19.2 & 79.6 & 39.7 & 30.4 & 47.6 & 49.3 \\
\rowcolor{black!7} Oracle distillation   & gold & 1.2 & 76.9 & 40.4 & 31.6 & 49.9 & 49.7 \\
\bottomrule
\end{tabular}}
\end{table*}
 
Table~\ref{tab:main} compares all methods trained on the same 83 unlabeled AMC prompts, evaluated transductively on AMC and, without further training, on three other benchmarks.

Among label-free methods, \canon{} is the strongest by a wide margin. It reaches 76.5 avg@32 on the transductive cell over a base of 66.0. The four RL-based alternatives cluster in a narrow band: TTRL at 70.3, EMPO at 69.6, SCRL at 71.0, and TTRL-Guard at 70.5. The two fine-tuning-based methods barely move the model: LMSI, which fine-tunes on consensus-filtered text with hard targets, reaches 66.5, and ScPO reaches 66.4, its vote-margin filter retaining only 12 preference pairs from the 83 prompts, too few to train on at test-time scale. The four RL methods differ in how they turn agreement into a scalar reward, from the raw majority vote to semantic-entropy minimization, consensus-gated positive and negative labels, and stability-weighted votes, yet they land within 1.4 points of one another and 5 to 7 points below \canon{}. What separates \canon{} is how much of the signal reaches the gradient. Strengthening LMSI and ScPO beyond their papers' protocols does not close the gap either (Appendix~\ref{app:impl}). Across three training seeds, \canon{}'s AMC result varies by 0.3 points, and across three evaluation seeds by 0.2 (Appendix~\ref{app:robustness}).

Consensus conditioning also recovers much of the value of gold conditioning. The oracle version, identical to \canon{} except that its teacher is anchored to deep-resampled gold-verified solutions, reaches 76.9 against \canon{}'s 76.5, a difference within evaluation noise. Trained transductively on AIME 2024 the two reach 48.0 and 44.6, and on GPQA 62.5 and 54.6 (Table~\ref{tab:ladder}); GPQA is the one benchmark where gold conditioning keeps a clear advantage (Appendix~\ref{app:significance}). Much of what a gold solution provides as privileged conditioning, the model's own consensus provides too.
 
Gold-reward reinforcement learning eventually surpasses \canon{} transductively, but at an order of magnitude more compute, and its extra gains do not transfer. GRPO reaches 79.6 at step 180, requiring 19.2 GPU-hours against 1.2 and gold labels throughout; even so, its remaining advantage of 3.1 points over \canon{} is not statistically significant under a paired bootstrap (see ~\ref{app:significance}). On the held-out columns the picture inverts: the step-180 GRPO checkpoint transfers no better than \canon{} on AIME 2024 (39.7 against 39.5), and worse on AIME 2025 (30.4 against 32.6) and GPQA (47.6 against 48.9). The additional transductive points that gold-reward RL buys with compute appear to be adaptation to the training prompts rather than transferable capability; we return to this in Section~\ref{subsec:transfer}.
 
Beyond the aggregate numbers, adaptation changes how the accuracy is delivered. After \canon{}, a single greedy pass on AMC scores 76.7 averaged over three evaluation seeds, up from 67.1 for the base model and approaching the base model's 32-sample majority vote of 83.1. Much of the benefit of ensembling is amortized into the weights, which is the practically relevant regime when per-query sampling budgets are constrained.

\begin{figure}[!h]
    \centering
    \begin{subfigure}[b]{0.525\textwidth}
        \centering
        \includegraphics[width=\linewidth]{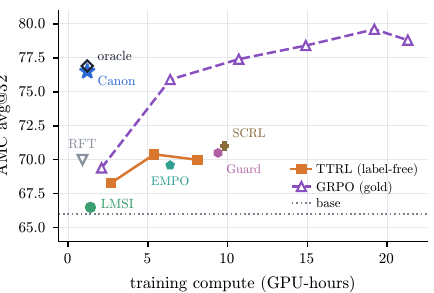}
        \caption{Accuracy against compute.}
        \label{fig:frontier}
    \end{subfigure}\hfill
    \begin{subfigure}[b]{0.455\textwidth}
        \centering
        \includegraphics[width=\linewidth]{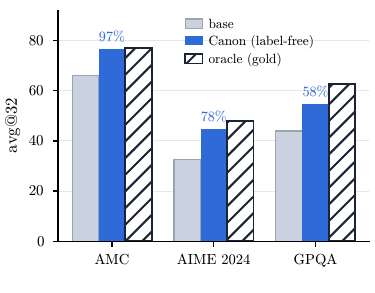}
        \caption{Oracle recovery.}
        \label{fig:recovery}
    \end{subfigure}
    \caption{\textbf{Transductive AMC, main comparisons} (Qwen3-4B-Instruct-2507). (a)~Accuracy against measured wall-clock on identical hardware, including generation. \canon{} reaches within half a point of the gold-conditioned oracle in 1.2 GPU-hours; gold-reward GRPO needs roughly 11 hours to match it and 19 to surpass it, and TTRL plateaus below \canon{} at every budget. Filled markers and solid lines denote label-free methods; open markers and dashed lines denote methods that use gold labels. (b)~\canon{} (label-free) against the deep-resampled oracle teacher (gold labels), per benchmark. The model's own consensus captures much of the gold solution's value as privileged context, with the largest remaining gap on GPQA; variance and paired bootstrap intervals in Appendix~\ref{app:significance}.}
    \label{fig:mainresults}
\end{figure}

\subsection{Breadth Across Benchmarks, Models, and Families (RQ2)}
\label{subsec:breadth}

\begin{table*}[!h]
\centering
\caption{\textbf{Transductive \canon{} across benchmarks, models, families, and scales} (avg@32). One fixed recipe per model, selected once on development slices and never on a reported benchmark (per-model values in Appendix~\ref{app:impl}); each cell trains only on that benchmark's unlabeled prompts. Deltas over the corresponding base model in parentheses. LFM2.5-8B-A1B is a mixture-of-experts (MoE) model with 1B active parameters. Majority-vote and coverage metrics for the Qwen3-4B-Instruct-2507 cells appear in Table~\ref{tab:capability}.}
\label{tab:breadth}
\small
\begin{tabular}{@{}l|ccccc@{}}
\toprule
 & \textbf{AMC} & \textbf{AIME24} & \textbf{AIME25} & \textbf{GPQA} & \textbf{MATH500} \\
\midrule
Qwen3-4B-Instruct-2507 & 66.0 & 32.5 & 30.1 & 43.9 & 89.3 \\
\;\; + \canon{} & \textbf{76.5} {\scriptsize\textcolor{teal}{(+10.5)}} & \textbf{44.6} {\scriptsize\textcolor{teal}{(+12.1)}} & \textbf{35.8} {\scriptsize\textcolor{teal}{(+5.7)}} & \textbf{54.6} {\scriptsize\textcolor{teal}{(+10.7)}} & \textbf{91.2} {\scriptsize\textcolor{teal}{(+1.8)}} \\
\midrule
Qwen3.5-2B & 36.8 & \phantom{0}8.2 & 11.8 & 27.1 & 68.8 \\
\;\; + \canon{} & \textbf{44.4} {\scriptsize\textcolor{teal}{(+7.6)}} & \textbf{14.7} {\scriptsize\textcolor{teal}{(+6.5)}} & \textbf{14.7} {\scriptsize\textcolor{teal}{(+2.9)}} & \textbf{35.2} {\scriptsize\textcolor{teal}{(+8.1)}} & \textbf{75.3} {\scriptsize\textcolor{teal}{(+6.6)}} \\
\midrule
Qwen3.5-4B & 67.6 & 37.4 & 26.2 & 52.5 & 88.6 \\
\;\; + \canon{} & \textbf{75.7} {\scriptsize\textcolor{teal}{(+8.1)}} & \textbf{43.4} {\scriptsize\textcolor{teal}{(+6.0)}} & \textbf{33.3} {\scriptsize\textcolor{teal}{(+7.2)}} & \textbf{61.6} {\scriptsize\textcolor{teal}{(+9.2)}} & \textbf{93.9} {\scriptsize\textcolor{teal}{(+5.3)}} \\
\midrule
Qwen3.5-9B & 74.5 & 43.1 & 34.3 & 59.2 & 91.9 \\
\;\; + \canon{} & \textbf{79.4} {\scriptsize\textcolor{teal}{(+4.9)}} & \textbf{49.4} {\scriptsize\textcolor{teal}{(+6.2)}} & \textbf{37.4} {\scriptsize\textcolor{teal}{(+3.1)}} & \textbf{68.2} {\scriptsize\textcolor{teal}{(+9.0)}} & \textbf{95.2} {\scriptsize\textcolor{teal}{(+3.3)}} \\
\midrule
SmolLM3-3B & 44.4 & \phantom{0}7.2 & 10.3 & 27.8 & 72.2 \\
\;\; + \canon{} & \textbf{49.0} {\scriptsize\textcolor{teal}{(+4.7)}} & \textbf{11.5} {\scriptsize\textcolor{teal}{(+4.3)}} & \textbf{11.9} {\scriptsize\textcolor{teal}{(+1.6)}} & \textbf{29.5} {\scriptsize\textcolor{teal}{(+1.7)}} & \textbf{77.8} {\scriptsize\textcolor{teal}{(+5.5)}} \\
\midrule
LFM2.5-8B-A1B (MoE, 1B active) & 58.6 & 28.8 & 23.3 & 23.0 & \textbf{83.1} \\
\;\; + \canon{} & \textbf{61.0} {\scriptsize\textcolor{teal}{(+2.4)}} & \textbf{31.9} {\scriptsize\textcolor{teal}{(+3.1)}} & \textbf{23.4} {\scriptsize\textcolor{teal}{(+0.1)}} & \textbf{30.6} {\scriptsize\textcolor{teal}{(+7.5)}} & 80.0 {\scriptsize\textcolor{gray}{($-$3.1)}} \\
\midrule
Gemma-4-E4B-it & 63.7 & 28.4 & 25.3 & 45.9 & 85.4 \\
\;\; + \canon{} & \textbf{72.1} {\scriptsize\textcolor{teal}{(+8.4)}} & \textbf{43.9} {\scriptsize\textcolor{teal}{(+15.4)}} & \textbf{28.3} {\scriptsize\textcolor{teal}{(+3.0)}} & \textbf{54.1} {\scriptsize\textcolor{teal}{(+8.2)}} & \textbf{91.1} {\scriptsize\textcolor{teal}{(+5.8)}} \\
\bottomrule
\end{tabular}
\end{table*}
 
Table~\ref{tab:breadth} applies \canon{} across five benchmarks and seven models spanning four families and scales from 2B to 9B, with one fixed recipe per model. The gains are broad: 34 of the 35 reported cells improve, by $+5.7$ to $+12.1$ points on the four hard benchmarks for Qwen3-4B-Instruct-2507 and by $+3.1$ to $+9.0$ for Qwen3.5-9B, the largest model. MATH500 shows that saturation is a property of the model-benchmark pair rather than the benchmark: Qwen3-4B-Instruct-2507 already answers 89\% of samples correctly and votes at 95\%, leaving little room between expressed and consensus accuracy, and gains $+1.8$, while the unsaturated models gain $+3.3$ to $+6.6$ on the same benchmark (Section~\ref{subsec:when}). AIME 2025 shows the smallest hard-benchmark gain for Qwen3-4B-Instruct-2507; its base consensus is also the weakest of the four (43.3 maj@32), again tracking consensus quality. Run transductively on AIME 2024 for comparison, TTRL reaches 39.3 avg@32 at its best checkpoint against \canon{}'s 44.6.

Evidence from a single model family can be weak on its own. \citet{shao2025spurious} show that reinforcement learning on Qwen models can produce large mathematical gains even from random or deliberately incorrect rewards, by amplifying reasoning behaviors already acquired during pretraining, that the same signals largely fail on other families, and accordingly recommend validating any proposed training signal across model families and against uninformative-signal controls. Our evidence addresses both recommendations. Within the Qwen family, the content-free conditioning control of Section~\ref{sec:ablations} plays the role of the uninformative signal and yields no gain, so the improvements require the consensus content. Across families, SmolLM3-3B improves on all five benchmarks ($+1.6$ to $+5.5$), and LFM2.5-8B-A1B, a mixture-of-experts model, improves on four of five, with the largest majority-vote gain of any reported cell on AMC ($+4.8$ points) and a 3.1-point decline on MATH500. Consensus-level effects otherwise vary by model: the Qwen3.5-9B and SmolLM3 majority votes are unchanged on AMC, and the 9B AMC cell is stable across three training seeds (mean 79.2, sd 0.7). Gemma-4-E4B-it, from a fourth family, improves on all five benchmarks, with the largest single-cell gain in the table on AIME 2024 ($+15.4$ points).
 
\subsection{Inductive Transfer and Self-Improvement (RQ3)}
\label{subsec:transfer}
 
The transductive setting cannot distinguish genuine improvement from adaptation to the particular prompts being evaluated. We therefore train label-free on one prompt set and evaluate on disjoint benchmarks, in two configurations: a single-source configuration, where the AMC-trained checkpoints of Table~\ref{tab:main} are evaluated elsewhere, and a pooled configuration, where training uses a larger mixed-mathematics pool.
 
The held-out columns of Table~\ref{tab:main} already contain the single-source result: AMC-trained \canon{} improves AIME 2024 by $+7.0$ points over base, AIME 2025 by $+2.5$, and GPQA by $+5.0$, the last a transfer from competition mathematics to graduate-level science questions. The AIME 2025 gain is small on its own; on the combined AIME 2024+2025 set ($n{=}60$), AMC-trained \canon{} improves by $+4.7$ points against $+2.7$ for TTRL and $+3.8$ for GRPO-180 (Appendix~\ref{app:significance}). On AIME 2026, a post-cutoff evaluation set constructed for this paper (Appendix~\ref{app:contamination}), AMC-trained \canon{} improves majority-vote accuracy from 51.7 to 58.6 with pass@1 moving within noise ($n{=}29$). Qwen3.5-4B shows the same pattern, including transfer from easy to hard problems within OmniMath (Appendix~\ref{app:transfer}).

\begin{table}[!h]
\begin{minipage}[c]{0.44\textwidth}
\centering
\caption{\textbf{Pooled label-free training, held-out evaluation.} Qwen3-4B-Instruct-2507 trained on a 339-prompt unlabeled pool (Appendix~\ref{app:pool}) and evaluated on AIME 2024, which shares no prompts with the pool. Trajectory methods (TTRL, GRPO) report their final checkpoint, since selecting an intermediate checkpoint would itself require labeled held-out evaluation; full trajectories in Figure~\ref{fig:pool}. Shaded rows use gold labels.
}
\label{tab:pool}
\small
\setlength{\tabcolsep}{4pt}
\begin{tabular}{llrcc}
\toprule
\textbf{Method} & \textbf{Labels} & \textbf{GPU-h} & \textbf{avg} & \textbf{maj} \\
\midrule
Base model    & ---  & 0    & 32.5 & 56.7 \\
TTRL-pool     & none & 34.5 & 39.4 & 63.3 \\
\canon{}-pool & none & 4.1  & \textbf{41.7} & 53.3 \\
\midrule
\rowcolor{black!7} GRPO-pool     & gold & 35.6 & 39.0 & 60.0 \\
\rowcolor{black!7} Oracle-pool   & gold & 4.1 & 42.4 & 56.7 \\
\bottomrule
\end{tabular}
\end{minipage}\hfill
\begin{minipage}[c]{0.53\textwidth}
\centering
\includegraphics[width=\linewidth]{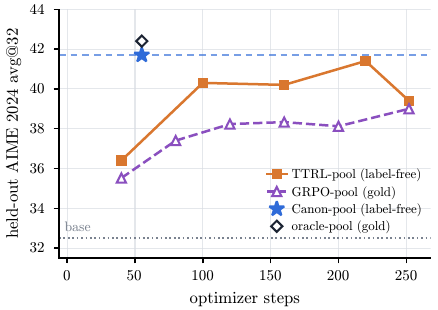}

\captionof{figure}{\textbf{Held-out AIME 2024 during pooled label-free training.} The RL-based methods climb slowly toward, and do not reach, the level \canon{} attains in a single epoch (dashed guide); TTRL regresses past its peak checkpoint. Filled markers and solid lines denote label-free methods; open markers and dashed lines denote methods that use gold labels.}
\label{fig:pool}
\end{minipage}
\end{table}

Table~\ref{tab:pool} scales the source data: all methods train on the 339-prompt unlabeled pool and are evaluated on held-out AIME 2024. \canon{} reaches 41.7 avg@32 in a single epoch, above the final checkpoints of both TTRL-pool (39.4 after 34.5 hours) and gold-reward GRPO-pool (39.0), and within a point of gold-conditioned oracle distillation (42.4); for scale, one \canon{}-pool epoch is roughly 55 optimizer steps at this batch size, while the horizontal axis of Figure~\ref{fig:pool} counts the RL methods' PPO steps. TTRL does pass through a stronger intermediate checkpoint (41.4 at step 220, Figure~\ref{fig:pool}) before regressing, but identifying that checkpoint requires labeled held-out evaluation that the label-free setting does not provide; \canon{} needs no such selection, since it trains for exactly one epoch. That a label-free method matches gold-supervised training on held-out data mirrors the single-source finding above: dense consensus distillation appears to extract more transferable improvement per unit of supervision than sparse gold rewards, at roughly a seventh of the compute. Two caveats apply: the evaluation set has 30 prompts, so one prompt moves maj@32 by 3.3 points, and \canon{}-pool's majority-vote accuracy lands below base here while GRPO-pool's does not. Section~\ref{subsec:sharpening} examines consensus-level effects where the statistics are adequate.

Transfer also degrades gracefully with distance from the training distribution. Training Qwen3.5-4B on one unlabeled OmniMath slice and evaluating the same checkpoint at increasing distance yields gains of $+9.2$ points held-in, $+4.9$ on harder problems from the same source, $+0.6$ on a different mathematics benchmark, and $+3.9$ cross-domain on GPQA (Appendix~\ref{app:transfer}). Label-free self-improvement is strongest near the training distribution but does not invert far from it.
 
\section{Analysis}
\label{sec:analysis}
 
\subsection{More than sharpening? (RQ4)}
\label{subsec:sharpening}
 
A natural concern with any self-training signal is that it sharpens the output distribution, concentrating probability on answers the model already produces, rather than improving what the model can solve~\citep{huang2024sharpening, lin2026ttrlguard}. Understood as concentration toward each prompt's existing answers, sharpening can move the finite-sample metrics in either direction, but two outcomes lie beyond it in expectation: it cannot make the majority answer correct on a prompt whose dominant answer was wrong, and it cannot systematically produce  correct samples on prompts where no sampled answer was correct. \canon{} does concentrate the distribution, as expected: mean vote share on AMC rises from 0.68 to 0.81. The question is whether concentration is all that happens.

\begin{table}[!h]
\begin{minipage}[t]{0.43\textwidth}
\centering
\caption{\textbf{Consensus and coverage effects} of transductive \canon{} (Qwen3-4B-Instruct-2507). Concentration toward existing answers cannot flip a wrong majority to a right one, and cannot cover a previously unsolved prompt.}
\label{tab:capability}
\small
\setlength{\tabcolsep}{3.5pt}
\begin{tabular}{l cc cc}
\toprule
& \multicolumn{2}{c}{\textbf{maj@32}} & \multicolumn{2}{c}{\textbf{pass@32}} \\
\cmidrule(lr){2-3}\cmidrule(lr){4-5}
 & base & \canon{} & base & \canon{} \\
\midrule
AMC      & 83.1 & \textbf{85.5} & 85.5 & \textbf{90.4} \\
AIME24   & 56.7 & \textbf{60.0} & 56.7 & \textbf{66.7} \\
AIME25   & 43.3 & \textbf{46.7} & 46.7 & \textbf{63.3} \\
GPQA     & 62.1 & \textbf{64.7} & \textbf{74.2} & 73.7 \\
MATH500  & \textbf{95.4} & 95.2 & \textbf{97.2} & 96.8 \\
\bottomrule
\end{tabular}
\end{minipage}\hfill
\begin{minipage}[t]{0.55\textwidth}
\centering
\captionof{figure}{\textbf{Where the gains land.} Change in avg@32 by base-model difficulty band (AMC, transductive). Gains concentrate on the hardest prompts, including a $+33.3$-point pass@32 gain on never-solved prompts (hatched).}
\label{fig:strata}
\includegraphics[width=\linewidth]{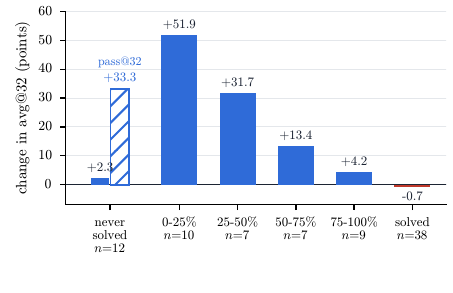}
\vspace{-1cm}
\end{minipage}
\end{table}
 
Concentration is not all of it.
Table~\ref{tab:capability} shows majority-vote accuracy improving on all four hard benchmarks, by $+2.4$ to $+3.3$ points, and pass@32, improving substantially on the mathematics benchmarks ($+4.8$ on AMC, $+10.0$ on AIME 2024, $+16.7$ on AIME 2025). The model's consensus becomes more often correct, and it solves problems it previously never solved in 32 attempts. Per-prompt transitions point the same way, with wrong-to-right majority flips outnumbering the reverse on every benchmark, though the per-benchmark counts are small (Appendix~\ref{app:predictors}).
Stratifying prompts by base difficulty makes the coverage effect explicit. Gains are largest in the hardest bands, $+51.9$ avg@32 points on prompts the base model solved in under a quarter of its samples and $+33.3$ pass@32 points on prompts it never solved, and are near zero on already-solved prompts (Figure~\ref{fig:strata}). GPQA and MATH500 show flat coverage, consistent with their headroom profiles discussed below. Further controls are provided in Appendix~\ref{app:predictors}; binning prompts on independent base-model evaluation draws preserves the difficulty profile, and on a fresh 64-sample draw \canon{}'s pass@$k$ stays above the base model's at every $k$, showing no sign of the pass@k crossover reported for RL-trained models \cite{yue2025rlvr}.

\subsection{When Does Consensus Supervision Help?}
\label{subsec:when}
 
The supervision \canon{} constructs is only as good as the consensus behind it, which predicts where the gains should concentrate. To test this, we pool per-prompt gains across twelve reported transductive cells of Tables~\ref{tab:main} and~\ref{tab:breadth}, giving 1{,}493 prompt-level observations across the three models, and classify each prompt by the state of the base model's consensus: whether the majority answer is correct, and whether it is held confidently (vote share of at least $0.5$). The prediction holds. Gains concentrate where the majority is correct but unsure ($+21.5$ points, 239 prompts); they are moderate where it is correct and saturated ($+4.3$, 875 prompts), small where it is wrong but weakly held ($+2.6$, 338), and negative where the samples collude on a wrong answer ($-3.0$, 41), the one regime in which the anchor points the wrong way. Confidently wrong consensus is also rare, which is why the aggregate effect is strongly positive.

\begin{table}[!h]
\centering
\caption{\textbf{Gains by base vote-share band}: change in avg@32 after transductive training, in fixed-width bins of the base model's vote share over its 32 evaluation samples, pooled over the twelve reported transductive cells.}
\label{tab:quintiles}
\small
\setlength{\tabcolsep}{6pt}
\begin{tabular}{lrrrrr}
\toprule
\textbf{Base vote share} & 0 to 0.2 & 0.2 to 0.4 & 0.4 to 0.6 & 0.6 to 0.8 & 0.8 to 1 \\
\midrule
$\Delta$avg@32 (points) & $+8.0$ & $+17.6$ & $+16.0$ & $+15.6$ & $+1.0$ \\
$n$ (prompts)           & 413    & 107     & 122     & 116     & 735 \\
\bottomrule
\end{tabular}
\end{table}

Table~\ref{tab:quintiles} shows the same pattern at finer grain. Gains are largest through the middle of the vote-share range, where the consensus is informative but not yet expressed in single samples; they are moderate at the bottom, where the consensus is often too weak to anchor reliably, and near zero at the saturated top, where there is nothing left to distill.
 
This yields a practical selection rule. \canon{} is indicated when base consensus accuracy meaningfully exceeds base pass@1, which is exactly the regime of small instruct models on competition mathematics and multiple-choice science, and contra-indicated where that gap has closed or the consensus is unreliable. Saturation is a property of the model-benchmark pair, not the benchmark: on MATH500 the gap is six points for Qwen3-4B-Instruct-2507, which gains $+1.8$, while the unsaturated Qwen3.5 models gain $+5.3$ and $+6.6$. The gap is measurable on a labeled development slice, and vote-share statistics alone provide a label-free proxy.
 
\section{Ablations}
\label{sec:ablations}
 
The gains of Section~\ref{sec:experiments} could in principle come from several sources: from the fine-tuning procedure itself, from conditioning on generic solution-shaped text, from the consensus answer leaking into the teacher, or from the reasoning content of the consensus solution. To separate these, we hold the entire recipe fixed and vary only what the teacher is conditioned on, from nothing at all up to the gold solution.
 
\begin{table}[!h]
\centering
\caption{\textbf{Conditioning ladder} (Qwen3-4B-Instruct-2507, transductive; avg@32 / maj@32). All arms are identical to \canon{} except for the teacher's conditioning content. Both oracles use deep-resampled gold solutions (protocol and coverage in Appendix~\ref{app:impl}).}
\label{tab:ladder}
\small
\begin{tabular}{ll|cc|cc}
\toprule
& & \multicolumn{2}{c|}{\textbf{AMC}} & \multicolumn{2}{c}{\textbf{GPQA}} \\
\textbf{Teacher conditioned on} & \textbf{Labels} & avg & maj & avg & maj \\
\midrule
--- (base, no training)          & ---  & 66.0 & 83.1 & 43.9 & 62.1 \\
Content-free placeholder         & none & 66.6 & 81.9 & 44.8 & 62.1 \\
Consensus final answer only      & none & 69.5 & 84.3 & 47.2 & 62.6 \\
Another prompt's consensus       & none & 69.9 & 83.1 & 47.7 & 64.1 \\
\textbf{Consensus solution (\canon{})} & none & \textbf{76.5} & \textbf{85.5} & \textbf{54.6} & \textbf{64.7} \\
Gold-verified solution (oracle)  & gold & 76.9 & 86.8 & 62.5 & 73.7 \\
\bottomrule
\end{tabular}
\end{table}
 
\begin{wraptable}{r}{0.445\textwidth}
\vspace{-10pt}
\centering
\caption{\textbf{Design and mechanism ablations} (transductive AMC, avg@32; one change at a time from the locked recipe). Teacher and rollout-freshness ablations and further detail in Appendix~\ref{app:design}.}
\label{tab:design}
\small
\setlength{\tabcolsep}{4pt}
\begin{tabular}{lc}
\toprule
\textbf{Variant} & \textbf{AMC} \\
\midrule
none (locked recipe)             & 76.5 \\
$K{=}3$ consensus solutions      & 75.6 \\
vote-share gate ($\tau{=}0.5$)   & 72.1 \\
majority-only supervision        & 73.0 \\
SFT on the consensus solution    & 67.4 \\
\bottomrule
\end{tabular}
\vspace{-8pt}
\end{wraptable}
Table~\ref{tab:ladder} reports the ladder. A content-free placeholder leaves the model essentially unchanged, ruling out the training procedure itself as the source of gains. Conditioning on the consensus answer alone recovers roughly a third of \canon{}'s improvement ($+3.5$ AMC, $+3.3$ GPQA over base), and conditioning on another prompt's consensus solution, correctly formatted but about the wrong problem, lands at the same level (69.9 and 47.7): solution-shaped text and a bare answer each explain only the first rung. The full consensus solution adds a further $+7.0$ and $+7.4$ on top. The reasoning trace about the problem at hand, not the answer token or the format, carries most of the signal, consistent with the view that the teacher's value lies in anchoring a line of reasoning rather than leaking a label.
 
The remaining design choices, and a set of mechanism arms that vary one component of the supervision at a time, are summarized in Table~\ref{tab:design} and detailed in Appendix~\ref{app:design}. Three findings carry the section. First, the teacher must be a frozen snapshot: a live shared-parameter teacher collapses training entirely, with generation entropy and the training loss reaching zero within a handful of steps (Appendix~\ref{app:design}). Second, the supervision must be dense and must cover all rollouts: sequence-level SFT toward the same consensus solution, selected by the same rule, gains $+1.4$ points where distillation gains $+10.5$, and masking out minority-answer rollouts costs $3.5$ points, so rollouts that disagree with the consensus still carry signal, consistent with the vote-share-gate ablation. Third, the budget knobs are forgiving: increasing the rollout count $N$ improves results monotonically up to $N{=}32$ and plateaus at 64, with even $N{=}4$ recovering two thirds of the gain, and additional epochs with regenerated rollouts add at most a point before reversing (Appendix~\ref{app:design}).
 
\section{Conclusion}
\label{sec:conclusion}
 
We showed that a model's own consensus can serve as privileged context for self-distillation, turning the signal behind majority voting into dense, token-level, label-free supervision. Instantiated with a frozen snapshot teacher and a single training epoch, \canon{} outperforms label-free reinforcement learning and consensus fine-tuning on transductive test-time training at a fraction of their compute, approaches gold-conditioned teachers, and transfers to held-out benchmarks, where it matches training methods that use gold labels. The gains go beyond pure sharpening: consensus accuracy and coverage rise alongside pass@1, concentrated on problems the base model rarely or never solved within its sampling budget.
 
\subsection{Limitations}
\label{subsec:limitations}
 
\canon{} requires an extractable final answer to form a consensus, restricting it to verifiable-output domains; replacing exact-match voting with minimum-Bayes-risk or medoid selection over sampled solutions would extend consensus extraction to free-form generation, and we leave this to future work. The method distills the gap between expressed and consensus accuracy, so it offers little on saturated model-benchmark pairs and can reinforce errors where the samples collude on a wrong answer, a measurable but rare failure mode (41 of 1493 pooled prompt-level observations in our analysis). We study a single round of self-distillation; in preliminary experiments, iterating the procedure yields at most one additional productive round before returns turn negative. Finally, the
transductive regime assumes access to test prompts before answering, which is natural for batch workloads but not for interactive use; the inductive results suggest the assumption can be relaxed.

\subsection{Acknowledgments}
The authors thank the Netherlands Organization for Scientific Research (NWO) for their support (VICI grant VI.C.212.053).
This work used the Dutch national e-infrastructure with the support of the SURF Cooperative using grant no. EINF-15485.
We acknowledge EuroHPC Joint Undertaking for awarding us access to Leonardo at CINECA, Italy.
 
\bibliographystyle{plainnat}
\bibliography{iclr2026_conference}
 
\clearpage
\appendix
 
\section{Formal Setup, Objective, and Training}
\label{app:formal}

This appendix restates the method of Section~\ref{sec:method} with full notation.

\textbf{Setup.} Let $\mathcal{V}$ be the vocabulary and $\mathcal{Y} = \mathcal{V}^{*}$ the space of token sequences. A policy $\pi_{\theta}(y \mid x) = \prod_{t=1}^{|y|} \pi_{\theta}(y_t \mid x, y_{<t})$ is initialized from a pretrained model $\pi_{0}$ and adapted through a low-rank parameterization of $\theta$. We are given unlabeled prompts $\mathcal{D} = \{x_1, \ldots, x_M\}$ and a deterministic answer-extraction map $\ans : \mathcal{Y} \to \mathcal{A} \cup \{\bot\}$, where $\mathcal{A}$ is the answer space (a canonical form for boxed expressions, or a choice letter) and $\bot$ marks sequences without an extractable answer. No reward function, verifier, or gold answer is available to the learner.

\textbf{Consensus.} For each prompt $x$, draw $N$ rollouts $y^{(1)}, \ldots, y^{(N)} \stackrel{\text{iid}}{\sim} \pi_{0}(\cdot \mid x)$ and let $a_i = \ans(y^{(i)})$. The majority answer is the plug-in mode of the induced answer distribution,
\begin{equation}
\hat{a}_x = \argmax_{a \in \mathcal{A}} \sum_{i=1}^{N} \mathbf{1}\big[a_i = a\big],
\end{equation}
with size ties resolved to the first-formed group and prompts whose rollouts are all unparseable removed from $\mathcal{D}$. Writing $\mathcal{S}_x = \{y^{(i)} : a_i = \hat{a}_x\}$ for the majority set, the consensus solution is its most confident member under the generating policy,
\begin{equation}
s_x = \argmax_{y \in \mathcal{S}_x} \; \frac{1}{|y|} \sum_{t=1}^{|y|} \log \pi_{0}\big(y_t \mid x, y_{<t}\big).
\end{equation}
Both selections are deterministic given the rollouts.

\textbf{Teacher.} Let $c(x, s)$ denote the augmented context that embeds a solution $s$ into the prompt for $x$ through the fixed template of Appendix~\ref{app:templates}. The teacher is the conditional next-token distribution of a frozen snapshot $\bar{\pi}$ of the policy under this context, $\bar{\pi}(\cdot \mid c(x, s_x), y_{<t})$. The snapshot is taken at the start of the epoch; in the single-epoch recipe, $\bar{\pi} = \pi_{0}$. The teacher is never sampled from: it is evaluated by teacher forcing on externally supplied token sequences, and no gradient flows through it.

\textbf{Objective.} Training minimizes the empirical token-mean Jensen-Shannon divergence between student and teacher next-token distributions over the sampled rollouts,
\begin{equation}
\label{eq:formalloss}
\hat{\mathcal{L}}(\theta) = \frac{1}{NM} \sum_{x \in \mathcal{D}} \sum_{i=1}^{N} \frac{1}{|y^{(i)}|} \sum_{t=1}^{|y^{(i)}|} \DJS{\pi_{\theta}\big(\cdot \mid x,\, y^{(i)}_{<t}\big)}{\bar{\pi}\big(\cdot \mid c(x, s_x),\, y^{(i)}_{<t}\big)},
\end{equation}
that is, each rollout contributes its token-mean divergence with equal per-sequence weight, so longer rollouts do not carry proportionally more gradient; $\DJS{p}{q} = \tfrac{1}{2}\DKL{p}{m} + \tfrac{1}{2}\DKL{q}{m}$ with $m = \tfrac{1}{2}(p + q)$, and both arguments are full distributions over $\mathcal{V}$, computed in the implementation as fp32 log-softmax over bf16 logits with no temperature. The divergence is symmetric, finite for any pair of distributions, and bounded by $\log 2$, which makes the per-token loss bounded regardless of how far teacher and student disagree; this contrasts with forward or reverse KL, either of which is unbounded under support mismatch. Treating the teacher as constant, the per-token gradient takes the simple form
\begin{equation}
\nabla_{\theta}\, \DJS{p_{\theta}}{q} = \frac{1}{2} \sum_{v \in \mathcal{V}} \nabla_{\theta}\, p_{\theta}(v)\, \log \frac{p_{\theta}(v)}{m(v)},
\end{equation}
which vanishes at the distributional optimum $p_{\theta} = q$. The log-ratio $\log \left(\frac{p_{\theta}(v)}{m(v)}\right)$ is bounded above by $\log 2$.

\textbf{Training.} One epoch of stochastic optimization over the fixed set of $\sum_x N$ sequences is performed with AdamW~\citep{AdamW} under the configuration of Appendix~\ref{app:impl}, with the teacher held fixed throughout. The rollouts serving as the distillation substrate are the same $N$ samples that defined the consensus, drawn from the policy at the start of the epoch; their staleness relative to the evolving student is therefore bounded by a single epoch of optimization. In the multi-epoch generalization, rollouts are regenerated and the teacher re-snapshotted at each epoch boundary, but all results in this paper use one epoch.

\section{Implementation Details}
\label{app:impl}
 
\paragraph{Training configuration.} \canon{} trains LoRA adapters (rank 64, $\alpha{=}128$, zero dropout, applied to all linear layers) with AdamW ($\beta_1{=}0.9$, $\beta_2{=}0.999$, weight decay 0.01), a constant learning rate of $10^{-5}$ with no warmup, gradient clipping at 1.0, and micro-batch size 1, for a single epoch over the $N{\times}M$ sampled sequences; on an 83-prompt benchmark with $N{=}32$ this is 2{,}656 sequences and roughly 13 optimizer steps. Rollout generation for training uses vLLM at temperature 1.0; the evaluation harness of Section~\ref{subsec:setup} (temperature 0.6) is never used to produce training rollouts. Training uses a response cap of 4{,}608 tokens (3{,}072 where memory-bound).
Training runs under FSDP via verl~\citep{sheng2024hybridflow} on four-GPU nodes; wall-clock figures are reported for a single reference node with four RTX 6000 Ada GPUs (48\,GB).
Each model's generation mode is fixed across training and evaluation (Table~\ref{tab:recipes}), multiple-choice training on GPQA uses a 2{,}048-token prompt cap, and answers are extracted from \verb|\boxed{}| with math-verify equivalence checking for mathematics and by letter extraction for multiple choice (Appendix~\ref{app:templates}).

\paragraph{Consensus mechanics.} In the consensus computation, vote share is measured over all $N$ rollouts including unparseable ones, and size ties between answer groups resolve to the first-formed group; the selection of the consensus solution by mean token log-probability is deterministic.

\paragraph{Software.} Training uses verl 0.9.0.dev0 with vLLM 0.22.1, transformers 5.12.1, torch 2.11.0, and math-verify 0.9.0; TTRL and GRPO instead run unmodified in TTRL's own released stack (verl 0.4.1, vLLM 0.8.4, transformers 4.52.4), so the faithful baseline keeps its reference environment.
 
\paragraph{Hyperparameter selection.} Because transductive experiments offer no validation set, hyperparameter selection must happen elsewhere. We therefore swept the learning rate once per model on a 64-prompt OmniMath development slice (difficulty 2 to 3.5), disjoint from every evaluation benchmark, and froze the selected value for all experiments in the paper. 
Table~\ref{tab:lrsweep} reports the sweep; $10^{-5}$ wins on avg@32 for both models and is used throughout, so no benchmark result in the paper reflects benchmark-specific tuning. The remaining models of Table~\ref{tab:breadth} follow the same protocol, with the learning rate, training response cap, and generation mode selected once per model on dedicated development slices and then frozen; Table~\ref{tab:recipes} lists the selected values.%

\begin{table}[!h]
\centering
\caption{Per-model recipe values (all other recipe components identical across models; everything selected on development slices).}
\label{tab:recipes}
\small
\begin{tabular}{@{}lccc@{}}
\toprule
\textbf{Model} & \textbf{LR} & \textbf{Train cap} & \textbf{Mode} \\
\midrule
Qwen3-4B-Instruct-2507 & $10^{-5}$ & 4{,}608 & non-thinking \\
Qwen3.5-4B / Qwen3.5-2B & $10^{-5}$ & 4{,}608 & non-thinking \\
Qwen3.5-9B & $10^{-5}$ & 3{,}072 & non-thinking \\
SmolLM3-3B & $3\cdot 10^{-5}$ & 4{,}608 & non-thinking \\
LFM2.5-8B-A1B & $5\cdot 10^{-6}$ & 4{,}608 & non-thinking \\
\bottomrule
\end{tabular}
\end{table}
 
\begin{table}[!h]
\centering
\caption{Development learning-rate sweep (OmniMath dev slice, avg@32 / maj@32). The selected value, in bold, is frozen for all reported experiments.}
\label{tab:lrsweep}
\small
\begin{tabular}{@{}l|ccc@{}}
\toprule
\textbf{Model} & $\mathbf{10^{-5}}$ & $5\cdot 10^{-6}$ & $2\cdot 10^{-6}$ \\
\midrule
Qwen3.5-4B & \textbf{73.4 / 81.3} & 71.6 / 82.8 & 70.3 / 82.8 \\
Qwen3.5-2B & \textbf{40.0 / 60.9} & 36.9 / 59.4 & 33.9 / 60.9 \\
\bottomrule
\end{tabular}
\end{table}
 
\paragraph{Baselines.} The baselines follow their reference implementations wherever one exists. TTRL is run faithfully from the released codebase and configuration (learning rate $5\cdot 10^{-7}$, GRPO advantage estimator, majority-vote pseudo-reward with 64 votes, 32 samples per prompt, response cap 3{,}072, no KL penalty), with training and validation both set to the unlabeled target prompts. GRPO is identical except that the reward verifies against gold answers, which makes the TTRL/GRPO pair a controlled comparison of pseudo-reward against gold reward under the same optimizer. LMSI follows the self-consistency-filtered SFT recipe~\citep{huang2022lmsi}: 32 rollouts per prompt are sampled at temperature 1.0, prompts whose majority vote share reaches 0.5 are kept (56 of 83 on AMC), and all majority-group rollouts become SFT targets (1{,}661 sequences), trained for one epoch at learning rate $10^{-5}$ with the same LoRA configuration; the reported row is a faithful retrain of this configuration whose transductive result matches the original run to 0.1 points, and its measured train pipeline (generation, consensus filtering, and SFT) takes 84 minutes end to end. RFT samples 16 rollouts per prompt at temperature $0.6$, keeps prompts with at least one gold-correct rollout (70 of 83 on AMC), and fine-tunes on the shortest correct rollout per prompt (LoRA rank 64, learning rate $10^{-5}$, 3 epochs, response-masked loss). ScPO is implemented from the paper, as no official code is released: 32 rollouts per prompt are clustered by final-answer equivalence with the repository's extraction machinery; the chosen response is a random member of the highest-vote cluster and the rejected response a random member of the least-voted cluster; prompts are kept only if the majority vote reaches $\tau k$ with $\tau{=}0.5$, and each pair is weighted by the vote margin. The objective is the paper's weighted DPO plus a weighted length-normalized NLL term ($\beta{=}0.5$, $\alpha{=}1$, learning rate $5 \cdot 10^{-6}$), with the frozen base model as reference via adapter disabling. Logged deviations from the paper: a single iteration and epoch, LoRA rank 64 rather than full fine-tuning, and a fresh seeded rollout draw. On the 83 AMC prompts the construction yields 12 pairs (44 prompts are unanimous and 27 fail the vote filter), which is itself an observation about consensus-preference methods at test-time scale. Strengthening these baselines beyond their papers' protocols does not change the picture: LMSI run for three epochs with fresh generation and re-filtering each epoch reaches 68.3, and a second ScPO iteration at the paper's iteration-two threshold ($\tau{=}0.7$) retains 3 pairs and reaches 66.0.

\paragraph{EMPO, SCRL, and TTRL-Guard.} All three run from their official releases on the same 83 prompts at the TTRL arm's budget: EMPO, which minimizes semantic entropy over answer clusters and serves as a confidence-sharpening representative~\citep{zhang2025empo}; SCRL~\citep{scrl2025};
and TTRL-Guard~\citep{lin2026ttrlguard}, whose released code reduces to its confidence-reweighting component in this setting, a caveat we log rather than repair. Their avg@32 trajectories at steps 20/40/60 are 68.2/70.0/69.6 (EMPO), 69.1/70.4/71.0 (SCRL), and 68.6/71.6/70.5 (TTRL-Guard): all end between 69.6 and 71.0, inside the TTRL band and well below \canon{}, and none exceeds the base model's majority-vote accuracy at its endpoint. Held-out, the three land in the TTRL band as well (avg@32 on AIME 2024 / AIME 2025 / GPQA: EMPO 36.9/31.7/45.0; SCRL 37.3/31.4/45.0; TTRL-Guard 34.9/33.2/45.7).

\paragraph{Oracle construction.} Oracle distillation is identical to \canon{} except that the teacher is conditioned on a gold-correct rollout, ranked by the same mean-token-log-probability statistic that selects the consensus solution. On AMC and AIME 2024 the conditioning solutions come from a dedicated deep-resampling pass rather than the training rollouts: up to 256 samples per prompt at temperature 1.0 under one seed, a second 256-sample pass at a second seed for prompts still unsolved, and 512 for the remainder, retaining the most confident gold-correct rollout per prompt. This yields gold conditioning for 75 of 83 AMC prompts, 23 of 30 AIME 2024 prompts, and 179 of 198 GPQA prompts, with per-prompt coverage records; the uncovered prompts (0 correct in 256 to 512 samples) fall back to consensus conditioning, and restricting the oracle comparisons to covered prompts changes no conclusion, since the fallback prompts are ones the base model never solves. The pooled oracle arm uses trainer-internal verification over the $N{=}32$ training rollouts instead. The resampling pass is a one-off cost excluded from the oracle's GPU-hours in Table~\ref{tab:main}.

\section{Prompt Templates and Answer Extraction}
\label{app:templates}

Every rollout, in training and evaluation, is generated from a single-turn chat prompt with no system message: the user message is the problem followed by the standard suffix used by TTRL, wrapped with the tokenizer's chat template (generation prompt appended; thinking disabled for hybrid-thinking models). The student computes its distributions under this same plain prompt.

\begin{promptbox}{Rollout and evaluation prompt (user message)}
{\small
\begin{verbatim}
{problem}
Please reason step by step, and put your final answer
within \boxed{}.
\end{verbatim}
}
\end{promptbox}

The teacher scores the identical rollout tokens under an augmented context: the consensus solution is inserted into a single user message that presents it as a reference solution, followed by the same assistant header, after which the rollout tokens are appended. The full teacher-side user message is the following (long lines wrapped for display; the reference solution sits inside the context, before the assistant turn, and the student never sees it):

\begin{promptbox}{Teacher conditioning prompt (user message)}
{\small
\begin{verbatim}
{problem}
Please reason step by step, and put your final answer
within \boxed{}.

A correct solution to this problem is given below for
your reference:
<solution>
{consensus solution}
</solution>

Guided by the reference solution, write your own
step-by-step solution, and put your final answer
within \boxed{}.
\end{verbatim}
}
\end{promptbox}

Final answers are read from the last \verb|\boxed{...}| occurrence in a response, with brace-depth matching for nested braces. For multiple choice, a cascade applies: a boxed letter (allowing \verb|\text| and \verb|\mathrm| wrappers), then an ``answer: X'' pattern, then a trailing bare letter, searched within the final 400 characters. Two answers are considered equivalent if they match as strings or if math-verify's symbolic check passes in either direction; responses without an extractable answer never match, and count in the vote-share denominator.

\section{The Pooled Training Set}
\label{app:pool}
 
The pooled experiments of Section~\ref{subsec:transfer} are meant to test whether label-free training on a broader unlabeled corpus improves a disjoint benchmark, so the pool must be large enough to matter and verifiably disjoint from the evaluation set. The pool contains 339 mathematics prompts: the 83 AMC 2023 prompts taken whole, a 128-prompt subset of MATH500 drawn uniformly at random (seed 1234, the project-wide seed), and 128 OmniMath problems drawn with the same seeded sampler from the difficulty 2.0 to 3.5 band used for development slices.
All prompts enter training unlabeled; gold answers for pool prompts are never read by any label-free method, and the label-full arms (GRPO-pool, oracle-pool) use them only as rewards or for teacher-rollout verification, exactly as in the single-source experiments.
 
The held-out evaluation set, AIME 2024, shares no prompts with the pool; a 13-gram overlap audit of all 339 pool prompts against every held-out evaluation set finds zero matches, with the AMC-transductive component as a positive control (Appendix~\ref{app:contamination}). One overlap inside the training side is disclosed for completeness: 43 of the pool's 128 OmniMath prompts also appear in the 128-prompt OmniMath development slice used for learning-rate selection. Both are training-side sets, so no evaluation benchmark is affected and the protocol claim of Section~\ref{subsec:setup} is unchanged.
 
\section{Significance Tests}
\label{app:significance}
 
The benchmark cells in this paper contain between 30 and 198 prompts, so point differences of a few points can be noise. We quantify uncertainty on the central comparisons with a paired bootstrap over prompts ($B{=}10{,}000$), reported in Table~\ref{tab:boot} as avg@32 differences in points with 95\% confidence intervals.
 
\begin{table}[!h]
\centering
\caption{Paired bootstrap comparisons (Qwen3-4B-Instruct-2507; $B{=}10{,}000$; avg@32 differences in points). All comparisons use the locked \canon{} artifact; transductive oracle comparisons use the deep-resampled oracle arms.}
\label{tab:boot}
\small
\setlength{\tabcolsep}{4pt}
\begin{tabular}{ll|rcc}
\toprule
\textbf{Comparison} & \textbf{Setting} & $\Delta$ & \textbf{95\% CI} & $n$ \\
\midrule
\canon{} $-$ base     & AMC transductive & $+10.5$ & $[+5.8, +15.6]$ & 83 \\
\canon{} $-$ TTRL60   & AMC transductive & $+6.2$  & $[+2.9, +9.9]$  & 83 \\
\canon{} $-$ LMSI     & AMC transductive & $+9.9$  & $[+5.5, +14.7]$ & 83 \\
\canon{} $-$ GRPO60   & AMC transductive & $+4.8$  & $[+0.8, +8.9]$  & 83 \\
\canon{} $-$ RFT      & AMC transductive & $+6.6$  & $[+1.8, +11.4]$ & 83 \\
Oracle $-$ \canon{}   & AMC transductive & $+0.4$  & $[-2.3, +3.4]$  & 83 \\
GRPO180 $-$ \canon{}  & AMC transductive & $+3.1$  & $[-0.3, +6.9]$  & 83 \\
\midrule
\canon{} $-$ base     & AIME24 transductive & $+12.1$ & $[+5.3, +19.8]$ & 30 \\
Oracle $-$ \canon{}   & AIME24 transductive & $+3.4$  & $[-1.4, +9.1]$  & 30 \\
Oracle $-$ \canon{}   & GPQA transductive & $+7.9$  & $[+4.7, +11.2]$ & 198 \\\midrule
\canon{}-AMC $-$ base & AIME24 held-out  & $+7.0$ & $[+3.3, +11.0]$ & 30 \\
\canon{}-AMC $-$ base & AIME25 held-out  & $+2.5$ & $[-1.9, +6.8]$  & 30 \\
\canon{}-AMC $-$ base & AIME24+25 combined & $+4.7$ & $[+2.0, +7.7]$ & 60 \\
\canon{}-AMC $-$ base & GPQA held-out    & $+5.0$ & $[+2.8, +7.2]$  & 198 \\
\canon{}-AMC $-$ TTRL-AMC & AIME24 held-out & $+2.8$ & $[+0.5, +5.7]$ & 30 \\
Oracle-AMC $-$ \canon{}-AMC & AIME24 held-out & $+0.9$ & $[-1.8, +3.9]$ & 30 \\
Oracle-AMC $-$ \canon{}-AMC & AIME25 held-out & $-1.0$ & $[-3.9, +2.0]$ & 30 \\
Oracle-AMC $-$ \canon{}-AMC & GPQA held-out  & $+1.1$ & $[-0.0, +2.2]$ & 198 \\
\midrule
\canon{}-pool $-$ base   & AIME24 held-out & $+9.2$ & $[+2.8, +16.7]$ & 30 \\
\canon{}-pool $-$ TTRL-pool & AIME24 held-out & $+2.3$ & $[-2.4, +9.2]$ & 30 \\
\canon{}-pool $-$ Oracle-pool & AIME24 held-out & $-0.7$ & $[-3.5, +1.9]$ & 30 \\
\bottomrule
\end{tabular}
\end{table}
 
Three conclusions follow. The advantage of \canon{} over the base model and over every label-free and matched-compute gold baseline is significant, with intervals clear of zero. Against the oracle the picture splits by domain: on the mathematics benchmarks the difference is not distinguishable from zero in any setting tested, while on GPQA the deep-resampled oracle keeps a significant transductive advantage and the held-out interval sits exactly at the boundary. Finally, GRPO-180's remaining transductive advantage, at sixteen times the compute and with gold labels throughout, is itself not significant.

\section{Design Choices and Variance}
\label{app:design}

The recipe of Section~\ref{subsec:canon} fixes four design choices: a frozen snapshot teacher, rollouts generated at the start of the epoch, a single consensus solution per prompt, and no vote-share gating. This appendix justifies each choice and reports seed variance; every experiment varies one choice at a time under an otherwise identical configuration.

\paragraph{Freezing the teacher.} Freezing is the one indispensable choice. Table~\ref{tab:teacher} compares the snapshot teacher against a live teacher that shares parameters with the student, on a 390-prompt OmniMath slice: the live-teacher run collapses entirely, and Table~\ref{tab:collapse} shows the collapse signature from the training diagnostics, with loss and gradient norm reaching exactly zero within five steps while every sampled response saturates the length cap with no extractable answer. Consensus anchoring is only stable against a fixed reference. Rollout freshness matters for the same on-policy reason, though far less dramatically: reusing stale rollouts from an earlier model version loses one to five points against regenerating at epoch start.

\begin{table}[!h]
\begin{minipage}[t]{0.44\textwidth}
\centering
\caption{\textbf{Teacher and rollout freshness} (390-prompt OmniMath slice, avg@32; one change at a time).}
\label{tab:teacher}
\small
\begin{tabular}{lc}
\toprule
\textbf{Variant} & \textbf{avg@32} \\
\midrule
frozen snapshot (recipe) & 72.1 \\
live-policy teacher      & 0.0 \\
stale rollouts           & 66.6 to 70.8 \\
\bottomrule
\end{tabular}
\end{minipage}\hfill
\begin{minipage}[t]{0.53\textwidth}
\centering
\caption{\textbf{Collapse signature} of the live-teacher run (training diagnostics at step 5) against the snapshot run.}
\label{tab:collapse}
\small
\setlength{\tabcolsep}{4pt}
\begin{tabular}{lcc}
\toprule
 & \textbf{Snapshot} & \textbf{Live teacher} \\
\midrule
training loss           & $\approx 0.15$ & 0.00 \\
generation entropy      & $\approx 0.28$ & 0.04 \\
gated prompts (frac.)   & 1.0            & 0.0 \\
responses at length cap & 0.11 to 0.34   & 1.00 \\
\bottomrule
\end{tabular}
\end{minipage}
\end{table}

\paragraph{Consensus width and gating.} The remaining two choices are matters of simplicity rather than necessity (values in Table~\ref{tab:design}). Using three consensus solutions instead of one, implemented as a uniform log-mixture of three consensus-anchored teacher distributions rather than concatenated text, costs about a point on AMC (76.5 to 75.6), so we keep one. Gating training on prompts with vote share of at least $0.5$ costs 4.4 points (76.5 to 72.1), consistent with the vote-share-band analysis of Section~\ref{subsec:when}: low-confidence consensus still carries usable signal, so discarding it wastes training data.

\paragraph{Rollout count.} $N$ controls both the consensus estimate and the amount of distillation substrate. Table~\ref{tab:nabl} varies $N$ with everything else locked: gains grow monotonically to $N{=}32$ and plateau at 64, and even $N{=}4$ recovers roughly two thirds of the full gain, so the method does not depend on a large sampling budget.

\paragraph{Additional rounds.} The recipe trains one epoch. Iterating the full procedure, i.e., drawing fresh rollouts and re-anchoring the teacher, buys little: on AMC a second round reaches 77.4 ($+0.9$, at twice the cost) and a third gives the gain back (76.4), and on the pooled configuration a second round declines on held-out AIME 2024 (38.8 against the single-round 41.7). One round captures the achievable gain.
\begin{table}[!h]
\centering
\caption{\textbf{Rollout-count ablation} (transductive AMC; locked recipe otherwise).}
\label{tab:nabl}
\small
\setlength{\tabcolsep}{7pt}
\begin{tabular}{lrrrrr}
\toprule
$N$ & 4 & 8 & 16 & \textbf{32 (recipe)} & 64 \\
\midrule
avg@32 & 73.0 & 73.5 & 75.3 & \textbf{76.5} & 76.1 \\
maj@32 & 83.1 & 83.1 & 84.3 & \textbf{85.5} & 85.5 \\
\bottomrule
\end{tabular}
\end{table}

\paragraph{Mechanism arms.} Two arms vary one component of the supervision at a time (Table~\ref{tab:design}). Restricting supervision to majority-answer rollouts (masking the minority) costs 3.5 points (76.5 to 73.0), so rollouts that disagree with the consensus carry usable signal, consistent with the gate ablation. Replacing distillation with sequence-level SFT toward the same consensus solution, selected by the same highest-confidence rule, yields 67.4 against \canon{}'s 76.5: the dense per-token target, not the choice of solution, carries most of the improvement. The divergence itself (JSD against forward KL) was selected on the OmniMath development slice, where JSD won; on the AMC test cell the divergence choice moves the result by under two points (forward KL 78.2, JSD 76.5), single cells at multi-seed noise of 0.3.

\paragraph{Stability.} Table~\ref{tab:seeds} reports three independent \canon{} training runs on AMC with identical configuration; three identical reruns of the AIME 2024 transductive cell give 46.0, 46.2, and 42.6 against the reported 44.6 (mean 44.8, sd 1.7), the expected larger variance of a 30-prompt set. Three seeds of the Qwen3.5-9B AMC cell give mean 79.2, sd 0.7. %
Evaluation-seed variance is measured separately in Appendix~\ref{app:robustness}.

\begin{table}[!h]
\centering
\caption{\textbf{Training-seed variance} (transductive AMC, three independent runs).}
\label{tab:seeds}
\small
\setlength{\tabcolsep}{7pt}
\begin{tabular}{lrrrr}
\toprule
 & seed 1 & seed 2 & seed 3 & mean (sd) \\
\midrule
avg@32 & 76.4 & 75.7 & 75.8 & 76.0 (0.3) \\
maj@32 & 86.8 & 84.3 & 85.5 & 85.5 (1.0) \\
\bottomrule
\end{tabular}
\end{table}

\section{Additional Transfer Results}
\label{app:transfer}
 
The main text establishes inductive transfer on Qwen3-4B-Instruct-2507. This appendix extends the evidence to Qwen3.5-4B and to two controlled notions of distance from the training distribution.

\begin{table}[!h]
\centering
\caption{\textbf{Additional inductive transfer on Qwen3.5-4B} (avg@32). Left: AMC-trained and pool-trained checkpoints evaluated on disjoint benchmarks. Right: one checkpoint trained label-free on a 128-prompt OmniMath slice, evaluated at increasing distance from the training distribution.}
\label{tab:transfer35}
\begin{subtable}[t]{0.45\textwidth}
\centering
\subcaption{Cross-benchmark transfer.}
\footnotesize
\setlength{\tabcolsep}{4pt}
\begin{tabular}{@{}lccc@{}}
\toprule
\textbf{Held-out} & \textbf{Base} & \textbf{AMC-tr.} & \textbf{Pool-tr.} \\
\midrule
AIME 2024 & 37.4 & 40.9 & 39.3 \\
AIME 2025 & 26.2 & 28.1 & 28.9 \\
GPQA      & 52.5 & 55.2  & 55.9 \\
\bottomrule
\end{tabular}
\end{subtable}\hfill
\begin{subtable}[t]{0.52\textwidth}
\centering
\subcaption{Distance ladder.}
\label{tab:distance}
\footnotesize
\setlength{\tabcolsep}{4pt}
\begin{tabular}{@{}lccc@{}}
\toprule
\textbf{Evaluation set} & \textbf{Base} & \textbf{Trained} & $\Delta$ \\
\midrule
Held-in slice (transductive) & 65.0 & 74.2 & $+9.2$ \\
Harder, same source          & 59.6 & 64.5 & $+4.9$ \\
Different benchmark (AMC)    & 67.6 & 68.2 & $+0.6$ \\
Cross-domain (GPQA)          & 52.5 & 56.4 & $+3.9$ \\
\bottomrule
\end{tabular}
\end{subtable}
\end{table}
 
Table~\ref{tab:transfer35}a shows the same qualitative picture as  Qwen3-4B-Instruct-2507: label-free training on AMC or on the pool improves every held-out benchmark tested, by two to three and a half points. Majority-vote accuracy on the 30-prompt AIME sets is noisy at this scale; it rises on AIME 2024 (from 60.0 to 63.3 for the AMC-trained checkpoint) and falls on AIME 2025 (from 43.3 to 36.7, a two-prompt movement), so we do not read consensus-level conclusions from these cells. Transfer also crosses difficulty within a single source: training on an easy OmniMath slice (difficulty 2 to 3.5) and evaluating on a difficulty 4.0 to 5.5 subset improves avg@32 from 29.6 to 31.3 with maj@32 unchanged.

Table~\ref{tab:transfer35}b evaluates one checkpoint at increasing distance. Gains decay with distance but remain non-negative everywhere; majority-vote accuracy on the far transfers is noisier, and on AMC it falls from 86.8 to 79.5 while avg@32 still improves. The pattern supports the reading of Section~\ref{subsec:transfer}: label-free self-improvement is strongest near the training distribution and attenuates, rather than inverts, away from it.

\section{Evaluation Robustness}
\label{app:robustness}

The benchmark cells in this paper are single evaluations under the frozen harness, so we quantify the sampling noise a reader needs to interpret point differences.

Table~\ref{tab:evalnoise} re-evaluates fixed checkpoints under two additional vLLM sampling seeds. The standard deviation over three seeds is 0.2 to 0.8 points on the 83-prompt AMC cells and 0.7 to 1.2 points on the 30-prompt AIME cells; differences of a point or less between arms on a 30-prompt set are therefore within evaluation noise, which is how we read them throughout. Training-seed variance for \canon{} is of the same order (0.3 points over three independent runs, Appendix~\ref{app:design}).

\begin{table}[!h]
\centering
\caption{\textbf{Evaluation-seed variance} (avg@32 under three vLLM sampling seeds; fixed checkpoints, frozen harness otherwise).}
\label{tab:evalnoise}
\small
\begin{tabular}{llcccc}
\toprule
\textbf{Checkpoint} & \textbf{Eval set} & \textbf{1234} & \textbf{5678} & \textbf{9012} & \textbf{sd} \\
\midrule
base            & AMC       & 66.0 & 67.2 & 65.7 & 0.8 \\
\canon{}        & AMC       & 76.5 & 76.9 & 76.7 & 0.2 \\
base            & AIME 2024 & 32.5 & 33.5 & 34.0 & 0.8 \\
\canon{}        & AIME 2024 & 44.6 & 46.5 & 44.2 & 1.2 \\
\canon{}-pool   & AIME 2024 & 41.7 & 40.3 & 41.0 & 0.7 \\
\bottomrule
\end{tabular}
\end{table}

Finally, re-evaluating every AMC arm under a stricter 3{,}072-token generation cap preserves the method ordering: base 57.3, TTRL 61.5, \canon{} 69.2, GRPO-180 74.9.

\section{Contamination Audits and a Post-Cutoff Evaluation}
\label{app:contamination}

This appendix addresses the possibility that inductive transfer reflects overlap between the training pool and the held-out benchmarks.

At the prompt level, an exact-string and normalized 13-gram overlap audit between all 339 pool prompts and each held-out evaluation set (AIME 2024, AIME 2025, GPQA) finds zero matches; the audit's positive control, AMC prompts against the pool's AMC component, recovers all 83 with near-duplicate pairs. A sentence-embedding audit (all-MiniLM-L6-v2~\citep{wang2020minilm}) over all pool-eval pairs finds no pair with cosine similarity of 0.70 or above (maximum 0.687, median top-1 similarity 0.51).

Finally, we constructed a post-cutoff evaluation set from the 2026 AIME, which post-dates the training data of every model in this paper. The set contains 29 of the 30 problems: every answer was cross-checked against at least two independent sources, and one problem was excluded because its published statement is genuinely disputed. The set has zero 13-gram overlap with the training pool. Table~\ref{tab:aime2026} evaluates the transfer checkpoints on it. At $n{=}29$, single-sample accuracy is within evaluation noise for every arm; majority-vote accuracy improves under the AMC-trained checkpoint ($+6.9$ points) and the oracle-pool checkpoint ($+6.9$). We report this cell primarily as a contamination control: a benchmark that cannot appear in any training corpus shows the same qualitative consensus-level movement as the pre-cutoff held-out sets.

\begin{table}[!h]
\centering
\caption{\textbf{Post-cutoff evaluation: AIME 2026} ($n{=}29$; avg@32 / maj@32). Checkpoints as in Section~\ref{subsec:transfer}.}
\label{tab:aime2026}
\small
\begin{tabular}{lcc}
\toprule
\textbf{Arm} & \textbf{avg@32} & \textbf{maj@32} \\
\midrule
Base model    & 36.1 & 51.7 \\
\canon{}-AMC  & 37.1 & 58.6 \\
\canon{}-pool & 35.9 & 55.2 \\
Oracle-pool   & 36.9 & 58.6 \\
\bottomrule
\end{tabular}
\end{table}

\section{Per-Prompt Analysis Details}
\label{app:predictors}

This appendix defines the per-prompt quantities used in Section~\ref{sec:analysis} and reports the statistics that did not fit in the main text. For each prompt we compare base and trained accuracy over the 32 evaluation samples and classify the movement. Coverage transitions count prompts that move between zero and at least one correct sample (cov$^{+}$ for gained coverage, cov$^{-}$ for lost); sharpening transitions count accuracy movement within already-covered prompts (sharp$^{+}$ and sharp$^{-}$); and majority flips compare the correctness of the per-prompt majority answer before and after training.

\begin{table}[!h]
\centering
\caption{\textbf{Per-prompt transitions} under transductive \canon{} (Qwen3-4B-Instruct-2507). Counts of prompts per category; the full per-cell table over all reported checkpoints is released with the evaluation artifacts.}

\label{tab:transitions}
\small
\begin{tabular}{@{}lccccc|cc@{}}
\toprule
& & & & & & \multicolumn{2}{c}{\textbf{maj flips to}} \\
& cov$^{+}$ & cov$^{-}$ & sharp$^{+}$ & sharp$^{-}$ & flat & right & wrong \\
\midrule
AMC (83)        & 4  & 0  & 26 & 9  & 44 & 3  & 1 \\
AIME 2024 (30)  & 4  & 1  & 11 & 2  & 12 & 3  & 2 \\
GPQA (198)      & 15 & 16 & 78 & 14 & 75 & 13 & 8 \\
\bottomrule
\end{tabular}
\end{table}

Table~\ref{tab:transitions} shows the transition counts on the three transductive benchmarks. On AMC the picture is clean: coverage only grows, sharpening improvements outnumber regressions roughly three to one, and majority flips run three to the right against one to the wrong side. AIME 2024 shows the same asymmetry at its smaller scale. On GPQA the aggregate is similar but individual prompts are noisier, with coverage gains and losses nearly balanced and majority flips at 13 right against 8 wrong; this matches the flat pass@32 and positive maj@32 of Table~\ref{tab:capability}.

\paragraph{A regression-to-the-mean control for the difficulty analysis.} The difficulty bands of Figure~\ref{fig:strata} are assigned from the base model's observed accuracy over 32 samples, and the same finite-sample estimate ordinarily serves as the baseline when computing gains. Prompts that enter the hardest bands partly through an unlucky draw would then show spurious gains under any re-evaluation, and the never-solved band can only move up. To control for this, we separate the roles: bands are assigned from one base evaluation draw, the baseline is computed from an independent second base draw, and the post-training accuracy comes from the trained checkpoint (the two additional evaluation seeds of Appendix~\ref{app:robustness}). Table~\ref{tab:rtm} reports both estimates. The regression component is small in every band except one, where it works against the naive estimate, and the aggregate gain is slightly larger under the control ($+11.2$ against $+9.7$ points). On the never-solved band, the independent draw also scores zero. Band membership and magnitudes differ from Figure~\ref{fig:strata} because the draws differ.

\begin{table}[!h]
\centering
\caption{\textbf{Regression-to-the-mean control} (AMC, transductive \canon{}, avg@32 in percent). Bands assigned from base draw A; $\Delta$naive uses draw A as the baseline, $\Delta$clean uses the independent base draw B.}
\label{tab:rtm}
\small
\setlength{\tabcolsep}{5pt}
\begin{tabular}{lrrrrrr}
\toprule
\textbf{Band (by draw A)} & $n$ & \textbf{base A} & \textbf{base B} & \textbf{\canon{}} & $\Delta$\textbf{naive} & $\Delta$\textbf{clean} \\
\midrule
never solved      & 10 & 0.0  & 0.0  & 2.5  & $+2.5$  & $+2.5$ \\
(0, 0.25]         & 11 & 11.6 & 11.4 & 51.1 & $+39.5$ & $+39.8$ \\
(0.25, 0.5]       & 7  & 40.2 & 40.6 & 70.1 & $+29.9$ & $+29.5$ \\
(0.5, 0.75]       & 7  & 62.9 & 50.9 & 85.3 & $+22.3$ & $+34.4$ \\
(0.75, 1]         & 48 & 98.6 & 97.7 & 98.1 & $-0.5$  & $+0.5$ \\
\midrule
all prompts       & 83 & 67.2 & 65.7 & 76.9 & $+9.7$  & $+11.2$ \\
\bottomrule
\end{tabular}
\end{table}

\paragraph{Coverage at a larger sampling budget.} The coverage claims of Section~\ref{subsec:sharpening} rest on 32 samples per prompt. pass@$k$ analyses of RL-trained models report a crossover: the trained model wins at small $k$ but falls below the base model at large $k$, indicating a narrowed reasoning boundary~\citep{yue2025rlvr}. Table~\ref{tab:pass64} extends the budget with a fresh 64-sample draw per prompt. \canon{} stays above the base model at every $k$, with no crossover through $k{=}64$, and its majority-vote advantage over pass@1 roughly halves, quantifying how much of the voting ensemble is amortized into the weights. A second round of \canon{} sits below the first at every $k$, consistent with the rounds analysis of Appendix~\ref{app:design}. Budgets beyond 64 samples, where the crossover of \citet{yue2025rlvr} typically emerges, are left to future work.

\begin{table}[!h]
\centering
\caption{\textbf{pass@$k$ at a larger budget} (AMC, fresh 64-sample draw per prompt, so values differ slightly from the fixed-seed harness). maj is majority vote over the 64 samples.}
\label{tab:pass64}
\small
\setlength{\tabcolsep}{7pt}
\begin{tabular}{lccccc}
\toprule
 & $k{=}1$ & $k{=}8$ & $k{=}32$ & $k{=}64$ & \textbf{maj} \\
\midrule
Base model             & 66.0 & 82.3 & 86.1 & 86.7 & 84.3 \\
\canon{}               & \textbf{77.1} & \textbf{87.9} & \textbf{92.0} & \textbf{94.0} & \textbf{86.8} \\
\canon{}, second round & 75.7 & 86.3 & 89.2 & 90.4 & 85.5 \\
\bottomrule
\end{tabular}
\end{table}

\end{document}